\pdfoutput=1

\documentclass{ecai} 

\usepackage{latexsym}
\usepackage{amssymb}
\usepackage{amsmath}
\usepackage{amsthm}
\usepackage{booktabs}
\usepackage{enumitem}
\usepackage{graphicx}
\usepackage{bm}
\usepackage{multirow}

\usepackage[ruled, vlined, linesnumbered]{algorithm2e}
\usepackage{algpseudocode}
\algdef{SE}[SUBALG]{Indent}{EndIndent}{}{\algorithmicend\ }%
\algtext*{Indent}
\algtext*{EndIndent}
\algnewcommand\algorithmicforeach{\textbf{for each}}
\algdef{S}[FOR]{ForEach}[1]{\algorithmicforeach\ #1\ \algorithmicdo}
\SetKwComment{Comment}{$\triangleright$\ }{}

\DeclareMathOperator*{\argmax}{arg\,max}  

\newcommand\Tstrut{\rule{0pt}{2.6ex}}         
\newcommand\Bstrut{\rule[-0.9ex]{0pt}{0pt}}   
\def\thickhline{\noalign{\hrule height.8pt}}

\newcommand{\BibTeX}{B\kern-.05em{\sc i\kern-.025em b}\kern-.08em\TeX}

\makeatletter
\def\blfootnote{\xdef\@thefnmark{}\@footnotetext}
\makeatother

\begin{document}


\begin{frontmatter}


\paperid{841} 


\title{Take a Step and Reconsider: Sequence Decoding \\ for Self-Improved Neural Combinatorial Optimization}


\author[A,B]{\fnms{Jonathan}~\snm{Pirnay}}
\author[A,B]{\fnms{Dominik G.}~\snm{Grimm}\thanks{Corresponding Author. Email: dominik.grimm@\{hswt,tum\}.de}}

\address[A]{Technical University of Munich, TUM Campus Straubing}
\address[B]{University of Applied Sciences Weihenstephan-Triesdorf}


\begin{abstract}
The constructive approach within Neural Combinatorial Optimization (NCO) treats a combinatorial optimization problem as a finite Markov decision process, where solutions are built incrementally through a sequence of decisions guided by a neural policy network. To train the policy, recent research is shifting toward a 'self-improved' learning methodology that addresses the limitations of reinforcement learning and supervised approaches. Here, the policy is iteratively trained in a supervised manner, with solutions derived from the current policy serving as pseudo-labels. The way these solutions are obtained from the policy determines the quality of the pseudo-labels. In this paper, we present a simple and problem-independent sequence decoding method for self-improved learning based on sampling sequences without replacement. We incrementally follow the best solution found and repeat the sampling process from intermediate partial solutions. By modifying the policy to ignore previously sampled sequences, we force it to consider only unseen alternatives, thereby increasing solution diversity. Experimental results for the Traveling Salesman and Capacitated Vehicle Routing Problem demonstrate its strong performance. Furthermore, our method outperforms previous NCO approaches on the Job Shop Scheduling Problem.
\end{abstract}

\end{frontmatter}


\section{Introduction} \label{sec:introduction}

Combinatorial optimization (CO) problems, which are characterized by their discrete nature and often NP-hard complexity, are essential in many areas, including logistics, manufacturing, process design, and scheduling. Traditional methods often rely on heuristics that require domain expertise and struggle with scalability. In recent years, Neural Combinatorial Optimization (NCO) has emerged as a research area that aims to let deep neural networks learn to generate heuristics from the instance distribution of the problem at hand \citep{bengio2021machine}. Among various strategies within NCO, the {\em constructive} approach formulates a CO problem as a finite Markov decision process and represents a solution to an instance as a {\em sequence} of incremental decisions. A neural network computes a policy to guide these decisions.\blfootnote{Code available at: \url{https://github.com/grimmlab/step-and-reconsider}}

The policy network is typically trained with supervised learning (SL) techniques \citep{vinyals2015pointer,joshi2019efficient,fu2021generalize,kool2022deep,drakulic2023bq,luo2023neural} or reinforcement learning (RL) \citep{bello2016neural,kool2018attention,nazari2018reinforcement,hottung2020neural,kwon2020pomo,kim2021learning,park2021learning,wu2021learning,park2022schedulenet,zhang2020learning,hottung2022efficient,zhang2024deep}. Both approaches have particular challenges: SL-based methods require a large number of high-quality expert solutions to be used as labels, which are usually obtained from existing (exact) solvers. In the case of larger instances or complex problems, it can be challenging, if not infeasible, to pre-generate high-quality solutions. Conversely, RL-based methods do not necessitate the use of expert solutions, yet they are susceptible to the sparse reward problem \citep{hare2019dealing} and high hyperparameter sensitivity \citep{schulman2017proximal}. Furthermore, RL-based approaches typically use policy gradient methods (in particular, variants of REINFORCE \citep{williams1992simple}), where gradients are derived from complete trajectories, resulting in high computational cost. While state-of-the-art constructive RL-based approaches such as POMO \citep{kwon2020pomo} demonstrate remarkable performance on the training distribution, they exhibit limited generalizability to larger problem instances. To further complicate matters, it has recently been shown \citep{luo2023neural,drakulic2023bq} that the poor generalization is likely due to the lightweight decoder structure of commonly used architectures \cite{kool2018attention}. Instead, \citet{luo2023neural} and \citet{drakulic2023bq} suggest increasing the decoder size, which leads to significant memory requirements for policy gradient methods. 

To bridge SL- and RL-based methods, recent work \cite{luo2023neural,corsini2024self,pirnay2024self,luo2024self} diverges from RL to a more straightforward, {\em self-improved learning} (SIL) approach: During training, the current policy network is decoded (e.g., by sampling) to generate (or refine) solutions to randomly generated instances. The best solutions generated are used as pseudo-labels, which the network is trained to imitate via SL. Repeating this process creates a "self-improving" loop. As no gradients need to be collected during decoding, large architectures can be trained in this way. The challenge lies in identifying an effective decoding method that is {\bf (a)} capable of rapidly generating solutions for potentially thousands of instances that can be utilized in a single training epoch; {\bf (b)} able to provide diverse sequences for sufficient exploration but can be adapted to exploit the model in later training stages; and, {\bf (c)} generalizable across different problems, with few hyperparameters to adjust to the specific problem at hand. Given {\bf (a)}, time-consuming search methods such as Monte Carlo Tree Search (MCTS) are unsuitable. Although naive Monte Carlo i.i.d. sampling from the policy is theoretically sound \citep{corsini2024self}, its output is often not diverse. Furthermore, it can require a large number of samples (even at low annealing temperatures or Top-$p$/-$k$ sampling \cite{Holtzman2020The}) to let the model improve in later stages of training \citep{kool2019stochastic,shi2020incremental,kwon2020pomo,pirnay2024self}. On the other hand, sampling sequences {\em without replacement} (WOR) \cite{kool2019stochastic,shi2020incremental,meister2021conditional} yields diverse sequences. However, for CO problems, it has been observed that the advantage of sampling WOR over sampling with replacement diminishes with increasing solution length \cite{shi2020incremental,pirnay2024self}.  

In this paper, we propose a simple yet effective decoding mechanism for sequence models that exploits the diversity and parallelization capabilities of Stochastic Beam Search (SBS) \cite{kool2019stochastic} in an MCTS-like manner: We maintain a search tree where each node represents a partial solution and leaf nodes are complete solutions. Given a beam width $k$ and a step size $s$, we sample $k$ leaf nodes WOR from the model using SBS. We then remove the probability mass of the leaves from the tree, which marks the sequences as sampled. We select the best solution from the sampled leaves and, assuming it corresponds to a sequence $(a_1, \dots, a_n)$, follow it for $s$ steps. After shifting the tree's root to the partial solution $(a_1, \dots, a_s)$, we repeat the process of finding a better solution until we have traversed the entire tree.

The decoding method is fast, generalizable, and possesses only two intuitive hyperparameters, namely $k$ and $s$. These can be readily adapted to the available computational resources and problem length. By marking found sequences as sampled, we consistently consider {\em unseen} alternative solutions. By following the best solution for $s$ steps, we gradually reduce the length of the problem. This forces more diversity in the sampling process. 

We summarize our contributions as follows:
\begin{itemize}
	\item In the recent spirit of simplifying and scaling the training process of NCO methods, we propose a novel and straightforward sequence decoding method for SIL.
	\item We train two state-of-the-art architectures \cite{drakulic2023bq,luo2023neural} for the Traveling Salesman Problem (TSP) and the Capacitated Vehicle Routing Problem (CVRP) with 100 nodes in an SIL setting using our decoding method. Our method matches the performance of SL on expert trajectories when evaluated on the training distribution and shows similarly strong generalization performance on larger instances. 
	\item We further evaluate our method on the Job Shop Scheduling Problem (JSSP), consistently outperforming current state-of-the-art NCO methods. 
	\item We additionally show on various policies that our proposed decoding method significantly outperforms SBS-based sampling with the same computational budget.
\end{itemize}

\section{Related work}

\paragraph{Constructive NCO} The first application of neural networks to directly predict solutions to CO problems is attributed to the Pointer Network of \citet{vinyals2015pointer}. Originally trained via SL, \citet{bello2016neural} employ REINFORCE \cite{williams1992simple} with a learned value baseline. Since then, as in many other areas of deep learning, variants of the Transformer \cite{vaswani2017attention} have become the standard architecture choice for many NCO models \cite{kool2018attention,deudon2018learning,kwon2020pomo,lee2024attention,drakulic2023bq,luo2023neural,9776116}. To circumvent learning a value function, the policy networks are usually trained with self-critical policy gradient methods \cite{rennie2017self} over complete trajectories. In particular, POMO \cite{kwon2020pomo} exploits problem symmetries and samples solutions from every possible starting node for a single instance, thereby significantly diversifying the solutions found. POMO and similar RL-methods perform remarkably well on training distributions of up to 100 nodes in routing problems. However, they do not scale well to larger instance sizes. The recent methods BQ \cite{drakulic2023bq} and LEHD \cite{luo2023neural} attribute the poor generalization to the light decoder structure of the used Attention Model \cite{kool2018attention}.
In contrast, they propose significantly increasing the decoder size (e.g., up to nine transformer blocks in BQ). \citet{drakulic2023bq} and \citet{luo2023neural} train their models with SL on expert solutions to instances with only 100 nodes for routing problems and achieve state-of-the-art results when generalizing up to 1,000 nodes. However, the size of the architecture makes it challenging to train with policy gradient methods.

\paragraph{Self-improved learning} To overcome the difficulties associated with RL and SL for NCO, recent studies propose a "self-improving" training paradigm. The central concept during training is to use the current policy and generate solutions (i.e., sequences) to random instances, which are then used as pseudo-labels to train the network with SL in a next-token prediction setup. In the appendix of their LEHD paper, \citet{luo2023neural} describe a self-improvement method where the model is pre-trained with RL on small routing problems to be computationally feasible. The resulting model is used to generate solutions to a set of randomly generated larger problem instances. Exploiting problem symmetries, the solutions are further improved by re-unrolling the policy along random subtours. The policy is then trained to imitate the resulting set of solutions. With impressive results, \citet{luo2024self} further develop this approach for up to 100,000 nodes. However, the method is limited to routing problems where an optimal solution of a complete tour guarantees the optimality of any subtour. \citet{corsini2024self} propose a "self-labeling" strategy for the JSSP. They utilize vanilla Monte Carlo i.i.d. sampling with the current model during training to obtain increasingly optimal solutions for the model to imitate. \citet{pirnay2024self} employ a similar training strategy. They improve the sampling process by sampling solutions WOR over multiple rounds, with each round guiding the policy towards sequences that perform better than expected. This method is close to ours in the spirit of diversifying sequences by sampling WOR in multiple steps. However, their method requires scaling the advantages with problem- and training-dependent hyperparameters, which can be complex to tune.

\paragraph{Sequence decoding} There is a plethora of search methods at {\em inference time} to improve on the greedy output of a sequence model besides pure sampling or beam search, e.g. \cite{bello2016neural,hottung2022efficient,choo2022simulation,Holtzman2020The}. Related to our approach is Simulation-guided Beam Search \cite{choo2022simulation}, which performs beam search in an MCTS way by coupling the pruning step in beam search with greedy rollouts. However, due to its purely exploitative nature, it is an unsuitable choice for sequence decoding in the context of SIL. In general, AlphaZero-type algorithms \cite{silver2017mastering} share similarities with SIL. However, running multiple simulations in the MCTS for a single action choice is time-consuming and non-trivial to parallelize. Concerning sampling, considering a diverse set of solutions can significantly improve training \cite{kwon2020pomo,kim2021learning}. A prominent way of diversification is sampling WOR \cite{kool2019stochastic,shi2020incremental,meister2021conditional}. In this work, we use SBS \cite{kool2019stochastic}, as it can be parallelized in a manner analogous to regular beam search.

\section{Preliminaries}

\subsection{Problem formulation} We consider a CO problem with $n$ discrete decision variables. A solution to a problem instance is given by a tuple $(a_1, \dots, a_n)$, representing the $n$ variable assignments (we assume a given numerical order). Let $S$ be the space of all possible solutions $\bar a_{1:n} := (a_1, \dots, a_n)$. The goal is to find a solution that maximizes a pre-defined objective function $f \colon S \to \mathbb R \cup \{-\infty\}$. Here, $f$ maps infeasible solutions to $-\infty$.

The constructive approach formulates the problem autoregressively, where a value is chosen for $a_1$, then for $a_2$ given $a_1$, and so on, until a full solution $\bar a_{1:n}$ is created. The policy network $\pi_{\theta}$ to guide these incremental choices is a sequence model with parameters $\theta$. For a partial solution $\bar a_{1:d} = (a_1, \dots, a_d)$, with $d < n$, the policy $\pi_{\theta}$ computes the conditional distribution $\pi_{\theta}(a_{d+1} | \bar a_{1:d})$ over the choices $a_{d+1}$ for the $(d+1)$th decision variable. In particular, to obtain a complete solution from the policy, we begin with an empty tuple $\bar a_{1:0} := ()$ and autoregressively decode $a_d \sim \pi_\theta(\cdot | \bar a_{1:d-1})$. For a (partial) solution $\bar a_{1:d} = (a_1, \dots, a_d)$ with $d \leq n$, we denote by $\pi_\theta(\bar a_{1:d})$ the total probability $\pi_\theta(\bar a_{1:d}) = \prod_{i = 1}^d \pi_\theta(a_i | \bar a_{1:i-1})$.

To simplify the notation, we do not use explicit labels to distinguish problem instances. However, it will always be transparent in context to which instance a solution is to be assigned. We assume a constant problem length $n$ for simplicity, but our method applies equally to varying sequence lengths.

\subsection{Self-improved training cycle} \label{sec:sil_training_cycle}

\begin{figure}
  \centering
  \includegraphics[width=0.9\linewidth]{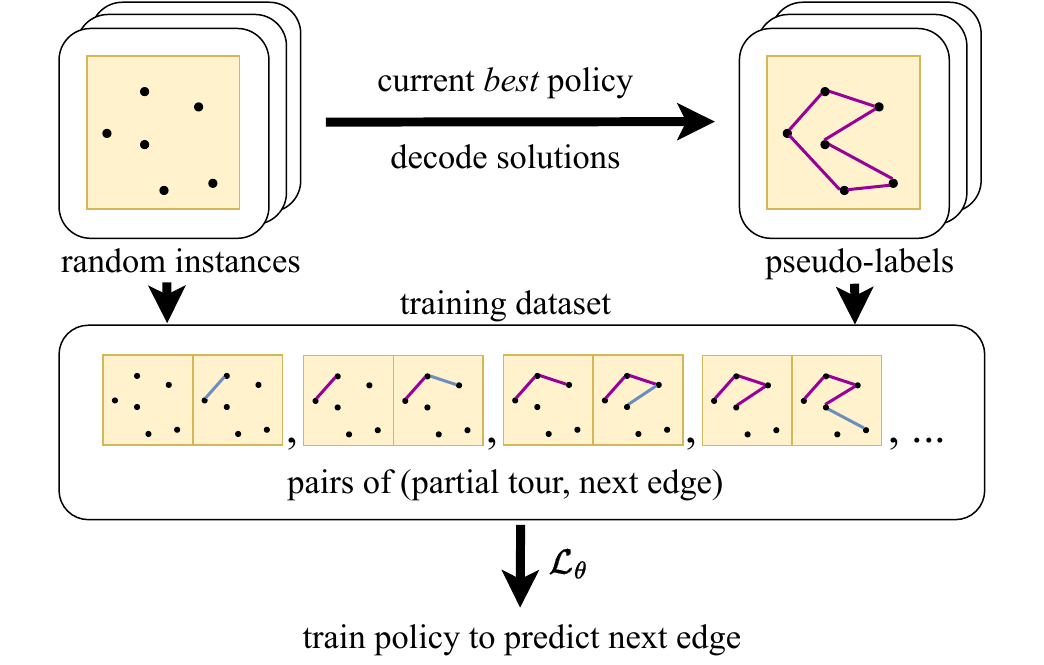}
  \caption{Overview of self-improved training with TSP as an illustrative example. A partial solution corresponds to a partial (unfinished) tour. The next sequence element for the model to predict is the next edge to be appended to the partial tour.}
  \label{fig:sil_overview}
  \vspace{20pt}
\end{figure}

The model $\pi_\theta$ is trained using the following simple SIL strategy, which we generalize from \cite{corsini2024self,pirnay2024self,luo2024self}. We present an illustration of the training cycle in Figure~\ref{sec:sil_training_cycle}. We randomly initialize parameters $\theta$ and keep best parameters $\theta'$. Initially, we set $\theta' \gets \theta$. We repeat the following steps for each training epoch:
\begin{enumerate}[label=(\arabic*)]
	\item Generate a set of random problem instances. For each instance, we use $\pi_{\theta'}$ to decode multiple solutions {\em in some way}, and keep the best solution found, according to the objective function $f$. The random instances and their best solutions found (used as pseudo-labels) build the training set for this epoch.
	\item Train $\pi_\theta$ on the generated data to predict the next sequence element from a partial solution with a cross-entropy loss: For a batch of size $B$ with solutions $\bar a_{1:n}^{j} = (a_1^j, \dots, a_n^j)$ for $j \in \{1, \dots, B\}$, we uniformly sample a partial solution $\bar a^j_{1:d_j}$ for each $j$ and $d_j < n$. The loss to be minimized is then given by
	\begin{align}
		\mathcal L_{\theta} = - \frac{1}{B}\sum_{j = 1}^B \log \pi_{\theta}\left(a_{d_j + 1}^j | \bar a_{1:d_j}^{j}\right).
	\end{align}
	\item At the end of the epoch, evaluate $\pi_\theta$ and $\pi_{\theta'}$ {\em greedily} on a fixed validation set. If $\pi_\theta$ outperforms $\pi_{\theta'}$, update the best parameters $\theta' \gets \theta$.
\end{enumerate}

In (1), for many problems, it is desirable to generate thousands of instances in each epoch (for comparison, the models in BQ \cite{drakulic2023bq} and LEHD \cite{luo2023neural} are trained with SL on 1M random instances and their optimal solutions). Consequently, the efficiency of the SIL strategy is strongly determined by the method used to decode the sequence model. 

\section{Method}

This section presents our main contribution, a sequence decoding method for efficient SIL. We also briefly recall SBS \cite{kool2019stochastic}, which forms the backbone of our method. In the following, we omit the parameters $\theta$ in the subscript of $\pi_\theta$.

\subsection{Sequence decoding as tree traversal}

As common for neural sequence models, we can view decoding $\pi$ for a problem instance as traversing a search tree from root to leaf. The root node corresponds to an empty sequence. A node in the tree at depth $d$ corresponds uniquely to a partial solution $\bar a_{1:d} = (a_1, \dots, a_d)$, and the direct children of this node represent the possible assignments to the $(d+1)$th decision variable. To be explicit, let $\text{Ch}(\bar a_{1:d})$ be the set of direct children of $\bar a_{1:d}$, then any $\bar b_{1:d+1} = (b_1, \dots, b_{d+1}) \in \text{Ch}(\bar a_{1:d})$ satisfies $b_i = a_i$ for $1 \leq i \leq d$. Thus, a leaf node corresponds uniquely to a complete solution.

Before explaining how we search the tree, we briefly describe how to {\em maintain} the tree.
When decoding the model for an instance, we create a search tree in memory and expand nodes as needed. When expanding a node $\bar a_{1:d}$, we query the model $\pi(\cdot | \bar a_{1:d})$ for the transition probabilities of its children. As described later, the transition probabilities will be modified during the search. Hence, for each node $\bar a_{1:d}$, we additionally keep an {\em unnormalized} total probability $p(\bar a_{1:d})$ which is set to the total probability $\pi(\bar a_{1:d})$ when the node is created. For a node $\bar a_{1:d}$ with parent $\bar a_{1:d-1}$, we denote by $\tilde \pi(a_d|\bar a_{1:d-1})$ the {\em normalized} transition probability
\begin{align}\label{eq:updated_seq_model}
	\tilde \pi(a_d|\bar a_{1:d-1}) = \frac{p(\bar a_{1:d})}{\sum_{\bar b_{1:d} \in \text{Ch}(\bar a_{1:d-1})}p(\bar b_{1:d})}.
\end{align}

\subsection{Stochastic Beam Search}

Ranking nodes by their total log-probability, a {\em beam search} of some beam width $k \in \mathbb N$ is a standard decoding method to obtain a set of $k$ unique high-probability sequences from the sequence model $\pi$. In beam search, all nodes within the current beam can be evaluated by $\pi$ in parallel, which aligns well with the effectiveness of GPUs on batches. Besides being classically a deterministic inference method, the $k$ sequences found with beam search often lack diversity. \citet{kool2019stochastic} present SBS, an elegant modification of beam search to sample $k$ sequences {\em without replacement} (WOR) from the sequence model $\pi$. The main idea is to perform regular beam search but perturb the total log-probability $\log \pi(\bar a_{1:d})$ by adding noise sampled from a standard Gumbel distribution. The Gumbel noise is sampled under the condition that the maximum perturbed log-probability of sibling nodes is equal to their parent's. This persists a node's perturbation down its subtree. 

The authors show that by sampling WOR from the distribution of complete sequences, SBS can obtain a set of sequences with high diversity. As it only changes the scoring of nodes, SBS can be implemented and, importantly, parallelized as a regular deterministic beam search.

\subsection{Take a step and reconsider}

\begin{figure*}
  \centering
  \includegraphics[width=0.80\linewidth]{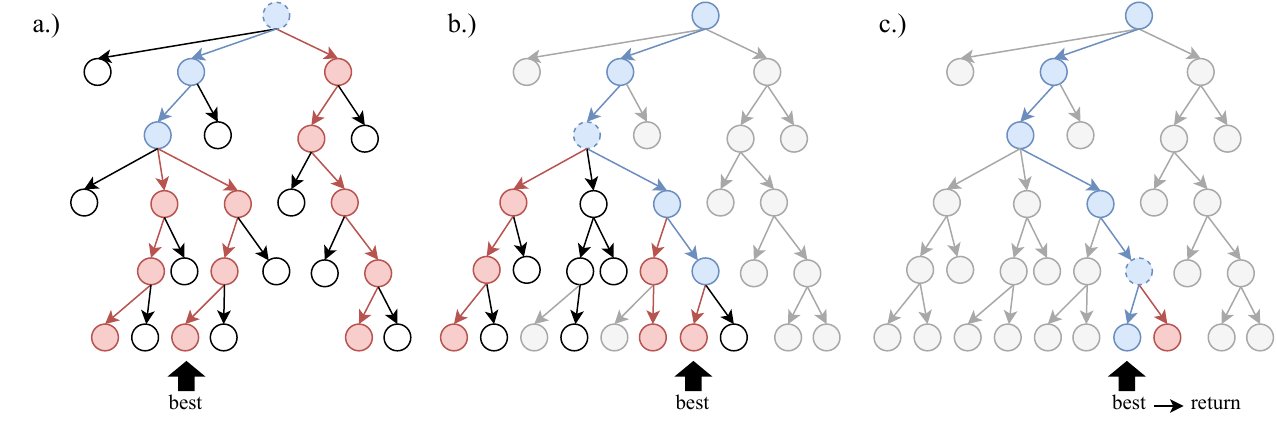}
  \caption{Example of sequence decoding with beam width $k=3$ and step size $s=2$. {\bf a.)} We sample $k$ leaves WOR (indicated in red) from the root node (dashed outline), creating nodes on demand. We follow the trajectory of the best solution for $s$ steps (indicated in blue). {\bf b.)} We shift the root node $s$, disregard the rest of the tree and remove the probability mass of sampled leaves (grayed out) from their ancestors. We sample $k$ unseen alternatives from the new root and find a better solution. We follow the new solution for $s$ steps. {\bf c.)} After shifting the root again, only one leaf is left to sample, which does not improve the current best solution.}
  \label{fig:method}
  \vspace{20pt}
\end{figure*}

We can now introduce our proposed decoding method. We summarize the method in Algorithm~\ref{algo:seq_decoding}, illustrate it in Figure~\ref{fig:method}, and give a brief walkthrough below.

\paragraph{Algorithm} 

\begin{algorithm}[t]
   \caption{Sequence decoding for SIL in NCO}
   \label{algo:seq_decoding}
      \DontPrintSemicolon
      \KwIn{$k \in \mathbb N$: beam width; $s \in \mathbb N$: step size}
      \KwIn{$\pi$: policy, $f$: objective function for problem instance}
      \BlankLine
      $\textsc{Root} \gets ()$ \Comment*[r]{root node}
      $t \gets 0$ \Comment*[r]{step count}
      $\bar b_{1:n} \gets \textsc{null}$ \Comment*[r]{best solution}
      \While{$t < n$}{
        $L = \{\bar a^1_{1:n}, \dots, \bar a^k_{1:n}\} \gets$ SBS with $\tilde \pi$ (see Eq. (\ref{eq:updated_seq_model})) and beam width $k$ from \textsc{Root}\;
        $\bar b_{1:n} \gets \argmax_{\bar a_{1:n} \in L \cup \{\bar b_{1:n}\}} f(\bar a_{1:n})$\;
        $t \gets \min\{t + s, n\}$\;
        \ForEach{$\bar a_{1:n} = (a_1, \dots, a_n) \in L$}{
         \If{$\bar a_{1:t} = \bar b_{1:t}$}{
         	\For{$i = t, \dots, n$}{
         		$p(\bar a_{1:i}) \gets p(\bar a_{1:i}) - \pi(\bar a_{1:n})$ \Comment*[r]{mark $\bar a_{1:n}$ as sampled in ancestor nodes}
         	}
         } 
        }
        $\textsc{Root} \gets \bar b_{1:t}$
      }
      \Return $\bar b_{1:n}$
\end{algorithm}

We require the choice of two hyperparameters: a beam width $k \in \mathbb N$ and a step size $s \in \mathbb N$ with $s \leq n$. Given a problem instance, we start with an empty tree and sample a set $L$ of $k$ leaves $L = \{\bar a^1_{1:n}, \dots, \bar a^k_{1:n}\}$ WOR using SBS with policy $\tilde \pi$ (which equals $\pi$ at the beginning). We take the best trajectory $\bar b_{1:n}$ sampled according to the objective function $f$. 
 We check for every sampled sequence $\bar a^j_{1:n} \in L$ if it shares its first $s$ steps with $\bar b_{1:n}$, i.e., if $(a^j_1, \dots, a^j_s) = (b_1, \dots, b_s)$. If so, we {\em mark it as sampled} by removing the leaf's probability $\pi(\bar a^j_{1:n})$ from its $i$-th ancestor's unnormalized probability mass $p(\bar a^j_{1:i})$, where $1 \leq i \leq n$. 

Now we follow $\bar b_{1:n}$ for $s$ steps and set $\bar b_{1:s}$ as the new root of the tree. We repeat the process above from the new root, noting that we use SBS with the {\em updated} sequence model $\tilde \pi$ (cf. (\ref{eq:updated_seq_model})) from which already seen sequences can no longer be sampled. We update the currently best sequence $\bar b_{1:n}$ if the newly sampled $k$ solutions contain a better one. Again, we mark the found sequences as sampled, follow the best solution for another $s$ steps, and so on until we have traversed the entire tree.

Note that for numerical stability, we work with log-probabilities in practice.

\paragraph{Exploring unseen alternatives} The critical thing to note is that by removing the probability mass of a leaf node from all its ancestors, sampling from the updated policy $\tilde \pi$ for SBS becomes equivalent to sampling WOR from $\pi$ {\em under the condition} that already encountered sequences can not be sampled again. This strategy was originally proposed by \cite{kool2020ancestral,shi2020incremental} to sample sequences WOR incrementally. By taking only a limited number of steps $s$ from the best solution found and sampling again, the model is forced to consider only unseen alternatives. Furthermore, the length of the sequences to sample gradually shrinks. This is beneficial when refining the best solution further down in the tree, as sampling WOR is much more potent (compared to sampling with replacement) on shorter sequences \cite{shi2020incremental,kool2019buy,pirnay2024self}.

\paragraph{Simplicity} One of the advantages of SIL approaches lies in the simplicity of its training cycle (see Section~\ref{sec:sil_training_cycle}), as it can foster reproducibility and adaption of the method. We aim to devise the sequence decoding - the heart of SIL - in the same spirit. The interpretation of the two hyperparameters is intuitive: $k$ is the number of sequences to consider before temporarily committing to a solution ('the more, the better'), and $s$ is how long to commit to a solution before exploring alternatives ('the shorter, the better'). Furthermore, $k$ and $s$ can be easily adjusted to available computational resources (or as training progresses), where $s=1$ leads to an MCTS-like search, and $s=n$ reduces to simple SBS with width $k$. 

\paragraph{For inference} We couple the sequence decoding with Top-$p$ (nucleus) sampling \cite{Holtzman2020The} to use it as an inferencing method for a trained policy. Here, the unreliable tail of the distribution is trimmed from $\tilde \pi$ in each expansion step of SBS. We evaluate its effectiveness in Section~\ref{sec:experiments}.

\section{Experiments} \label{sec:experiments}

We evaluate the efficiency of our sequence decoding method within the SIL framework (cf. Section~\ref{sec:sil_training_cycle}) on the Euclidean TSP and CVRP and the standard JSSP.

\paragraph{Code} Our code in PyTorch and models are available at \url{https://github.com/grimmlab/step-and-reconsider}. We perform training and evaluation using four NVIDIA GeForce RTX 3090 with 24GB RAM.

\subsection{Routing problems} 

\paragraph{Traveling Salesman Problem} A problem instance of the two-dimensional Euclidean TSP in the unit square is given by the coordinates of $N$ nodes $x_1, \dots, x_N \in [0,1]^2 \subseteq \mathbb R^2$. The goal is to find a {\em roundtrip} that minimizes the total tour length, where the distance between two nodes $x_i, x_j$ is given by the Euclidean norm $\|x_i - x_j\|_2$. A complete tour is constructed sequentially by choosing one unvisited node after another (see Figure~\ref{fig:sil_overview}).

\paragraph{Capacitated Vehicle Routing Problem} In the CVRP, a {\em delivery vehicle} of capacity $D \in \mathbb R_{>0}$ needs to visit $N$ customer nodes $x_1, \dots, x_N \in [0,1]^2 \subseteq \mathbb R^2$. Each customer $x_i$ has a demand $\delta_i \in \mathbb R_{>0}$, which must be fulfilled by the vehicle. A feasible solution is given by a set of subtours which all start and end at a given {\em depot node}, where all customers are visited, and the sum of customer demands satisfied by each subtour does not exceed the capacity $D$. The aim is to find a feasible solution with minimal total tour length. Following the standard constructive formulation \cite{kool2018attention,drakulic2023bq,luo2023neural}, visiting the depot is not seen as a separate step: a complete tour is constructed by deciding for each unvisited customer node whether it is reached via the depot or directly from the previous customer. This ensures solution alignment, as the length of two feasible solutions is the same, even if they contain a different number of subtours. Our problem setup is identical, so we refer to \cite{luo2023neural} for details.

\paragraph{Data generation and optimal solutions} We consider TSP and CVRP instances of size $N \in \{100, 200, 500\}$. We generate random problem instances in the standard way by sampling node coordinates uniformly from the unit square. Additionally, for the CVRP, demands are sampled uniformly from $\{1, \dots, 9\}$. The capacity of the delivery vehicle is set to 50, 80, and 100 for a corresponding number of nodes 100, 200, and 500. For both problems, we train only on instances of size $N=100$ and test generalization on the larger sizes. For comparison with SL, we generate a random training dataset of one million instances and a validation set of 10k instances. The test set consists of 10k instances for $N=100$ (same set used in \cite{kool2018attention} for the TSP and \cite{luo2023neural} for the CVRP) and 128 instances for $N \in \{200,500\}$ (same sets used in \cite{drakulic2023bq} for the TSP and \cite{luo2023neural} for the CVRP). For the TSP, we obtain optimal solutions from the Concorde solver \cite{Concorde} to use as labels in SL and compute optimality gaps. For the CVRP, (near) optimal solutions are obtained from HGS \cite{vidal2022hybrid}.

\paragraph{Policy network architecture} We evaluate our approach using two recent state-of-the-art architectures for routing problems: the BQ architecture by \citet{drakulic2023bq} and the LEHD architecture by \citet{luo2023neural}. Both architectures are based on the Transformer \cite{vaswani2017attention} with a heavy decoder structure. In the original works, the models obtain strong generalization results but are trained with SL on expert data as the large architectures are unsuitable for training with RL. For CVRP with BQ, we stick to the original setup of nine transformer blocks with 12 attention heads and a latent dimension of 192. For TSP with BQ, the number of layers is the same, but with eight attention heads and a latent dimension of 128. For CVRP and TSP with LEHD, we use six transformer blocks in the decoder with eight heads and a latent dimension of 128. Similar to BQ, we use ReZero normalization \cite{bachlechner2021rezero} also for LEHD, as we found the training to be more stable (compared to no normalization as suggested in the original paper). For BQ and LEHD, the hidden dimension of the feedforward network in a transformer block is set to 512.

\paragraph{Training} For the SIL training, we decode in each epoch solutions in parallel to 1,000 random instances. We use a beam width of $k=64$ and step size $s=10$. Generating the solutions takes about 2 minutes on our setup. Using the generated best solutions as pseudo-labels, we train the model on 1,000 batches of 1,024 uniformly sampled subtours as in \cite{drakulic2023bq}. We apply the same training structure to LEHD. We use the Adam \cite{kingma2014adam} optimizer with an initial learning rate of 2e-4, clipping gradients to unit norm. To evaluate the improvement of the model, we test the policy on the pre-generated validation set after each epoch. We train the policy until we see no improvement on the validation set for 50 epochs. The setup is the same for both routing problems. However, we found the generalization performance for the CVRP to improve noticeably when, after the regular training, finetuning the model on solutions where the policy was heavily exploited. To this end, we decode another 30k solutions with $k=256$ and $s=1$ and Top-$p$ sampling with $p=0.8$ and continue to train the model for another 100 epochs. For comparison with SL, we use the same training setup but sample batches from the pre-generated training set of 1M instances.  

In total, this amounts to training the BQ model with SIL for $\sim$3k epochs on the TSP ($\sim$4.5k with SL) and $\sim$3k epochs on the CVRP ($\sim$1.5k with SL). The LEHD model converges faster. With SIL, it is trained for $\sim$2k epochs on the TSP ($\sim$2k with SL) and $\sim$1k epochs on the CVRP ($\sim$1k with SL).

\paragraph{Baselines} The primary baselines are given by SL with the corresponding identical BQ or LEHD architecture. Furthermore, we include four common constructive NCO baselines, namely {\bf (a)} the widely used Attention Model (AM) with a beam search of width 1,024 \cite{kool2018attention}, {\bf (b)} its multi-decoder counterpart (MDAM) \cite{xin2021multi} with beam search of width 50, {\bf (c)} POMO \cite{kwon2020pomo}, the state-of-the-art constructive method on the training distribution, with their most potent inference technique, and {\bf (d)} Simulation-guided Beam Search (SGBS) \cite{choo2022simulation} with POMO backbone and parameters $(\beta, \gamma)$ set to $(10,10)$ for TSP and $(4,4)$ for CVRP. As comparison partners for SIL methods, we list the results (TSP only) of {\bf (e)} LEHD pre-trained with RL on small-scale instances and finetuned with SIL (LEHD RL+SIL) as reported in \cite{luo2023neural}, and {\bf (f)} the Gumbeldore (GD) training strategy (GD SIL (BQ resp. LEHD)) \cite{pirnay2024self}, where the sampling process is pushed toward regions with higher advantage.

\paragraph{Results} We summarize the results in Table~\ref{table:results:routing} and group them by the used architectures. Bold indicates the best optimality gap per group. Results for LEHD RL+SIL and GD SIL are taken from the original papers \cite{luo2023neural,pirnay2024self}. On the TSP, we obtain excellent greedy results that even outperform the SL counterpart on the training distribution of $N=100$, with similarly strong generalization capabilities. In particular, we outperform GD SIL, which has a more complex decoding strategy. The same dynamic can be observed on the CVRP with BQ, with worse but still strong generalization results. Our SIL method with LEHD is close to, but does not fully reach, the SL results for CVRP. We note a general gap of about 1\% between the LEHD SL results for CVRP in the original paper \cite{luo2023neural} and our reproduced results, which we attribute to the slightly different training method we aligned with BQ. At the bottom, we group the non-greedy results of our trained BQ model using beam search and our sequence decoding method as an inference technique (coupled with Top-$p$ sampling).

\begin{table*}[!t]
   \caption{Results for TSP and CVRP. 'bs' means beam search with a given width. The last two rows result from applying our proposed decoding method with Top-$p$ sampling ($p=0.95$ for TSP and $0.8$ for CVRP). LEHD RL+SIL and GD SIL (LEHD) only report results on the TSP. Gaps are obtained with respect to Concorde \citep{Concorde} for TSP and HGS \citep{vidal2022hybrid} for CVRP. Reported times are the duration of solving all instances.}
   \label{table:results:routing}
   \vspace{20pt}
   \begin{center}
     \begin{tabular}{l cc cc cc | cc cc cc}
     \Bstrut Method & \multicolumn{2}{c}{Test (10k inst.)} & \multicolumn{4}{c|}{Generalization (128 inst.)} & \multicolumn{2}{c}{Test (10k inst.)} & \multicolumn{4}{c}{Generalization (128 inst.)} \\
     \thickhline
      \Tstrut \Bstrut & \multicolumn{2}{c}{TSP $N=100$} & \multicolumn{2}{c}{TSP $N = 200$} & \multicolumn{2}{c|}{TSP $N=500$} & \multicolumn{2}{c}{CVRP $N=100$} & \multicolumn{2}{c}{CVRP $N=200$} & \multicolumn{2}{c}{CVRP $N=500$} \\
      & \Bstrut Gap & Time & Gap & Time & Gap & Time & Gap & Time & Gap & Time & Gap & Time \\
      \thickhline
     \Tstrut AM, bs1024 \cite{kool2018attention} & 2.49\% & 5m & 6.18\% & 15s & 17.98\% & 2m & 4.20\% & 10m & 8.18\% & 24s & 18.01\% & 3m \\
     POMO, augx8 \cite{kwon2020pomo} & 0.14\% & 15s & 1.57\% & 2s & 20.18\% & 16s & 0.69\% & 25s & 4.87\% & 3s & 19.90\% & 24s \\
     SGBS \cite{choo2022simulation} & {\bf 0.06\%} & 4m & {\bf 0.67\%} & 14s & 11.42\% & 5m & {\bf 0.08\%} & 7m & {\bf 2.58\%} & 20s & 15.34\% & 6m \\
     MDAM, bs50 \cite{xin2021multi} \Bstrut & 0.40\% & 20m & 2.04\% & 3m & {\bf 9.88\%} & 11m & 2.21\% & 25m & 4.30\% & 3m & {\bf 10.50\%} & 12m \\
     \hline
     BQ SL, greedy \cite{drakulic2023bq} \Tstrut & 0.40\% & 30s & {\bf 0.60\%} & 3s & {\bf 0.98\%} & 16s & 3.03\% & 50s & {\bf 2.63\%} & 4s & {\bf 3.75\%} & 22s \\
     GD SIL (BQ), greedy \cite{pirnay2024self} & 0.41\% & 30s & 0.64\% & 3s & 1.12\% & 16s & 3.26\% & 50s & 3.05\% & 4s & 3.89\% & 22s \\
     {\bf Ours (BQ)}, greedy \Bstrut & {\bf 0.37\%} & 30s & {\bf 0.60\%} & 3s & 1.10\% & 16s & {\bf 2.96\%} & 50s & 3.27\% & 4s & 5.77\% & 22s \\
     \hline
     LEHD SL, greedy \cite{luo2023neural} \Tstrut & 0.58\% & 25s & 0.95\% & 2s & 1.72\% & 11s & {\bf 4.26\%} & 40s & {\bf 3.77\%} & 6s & {\bf 4.36\%} & 12s \\
     LEHD RL+SIL, greedy \cite{luo2023neural} & 1.07\% & 25s & 1.45\% & 2s & 2.58\% & 11s & - & - & - & - & - & - \\
     GD SIL (LEHD), greedy \cite{pirnay2024self} & 0.40\% & 25s & 0.72\% & 2s & 1.43\% & 11s & - & - & - & - & - & - \\
     {\bf Ours (LEHD)}, greedy \Bstrut & {\bf 0.39\%} & 25s & {\bf 0.66\%} & 2s & {\bf 1.40\%} & 11s & 5.08\% & 40s & 4.60\% & 6s & 5.51\% & 12s \\
     \hline
     {\bf Ours (BQ)}, bs64 \Tstrut & 0.004\% & 7m & 0.04\% & 45s & 0.33\% & 3m & 1.07\% & 10m & 1.48\% & 50s & 3.40\% & 3m   \\
     {\bf Ours (BQ)}, $k=64, s=25$ & 0.003\% & 10m & 0.04\% & 70s & 0.22\% & 12m & 0.42\% & 25m & 0.62\% & 2m & 2.69\% & 20m  \\
     {\bf Ours (BQ)}, $k=128, s=10$ \Bstrut & {\bf 0.0009\%} & 35m & {\bf 0.02\%} & 4m & {\bf 0.18\%} & 50m & {\bf 0.14\%} & 50m & {\bf 0.27\%} & 6m & {\bf 2.19\%} & 50m \\
     \thickhline
     \end{tabular}
   \end{center}
\end{table*}

\subsection{Job Shop Scheduling Problem} 

\paragraph{Problem setup} The standard JSSP of size $J \times M$ is a CO problem with $J$ jobs, each consisting of $M$ operations with given processing times. Each job operation must run on exactly one of $M$ machines (precedence constraint), which are assigned to the operations bijectively. A machine can only process one operation at a time. The operations of a job must run in order. The objective is to find a feasible schedule that processes all operations of all jobs and has a minimum makespan. We follow the constructive formulation \cite{corsini2024self,pirnay2024self} where an unfinished job is chosen of which to schedule the next ready operation at each iterative step. In particular, a feasible solution can be represented by a (not necessarily unique) sequence of jobs.

\paragraph{Data} Random instances are generated in the standard way by uniformly sampling integer processing times from $[1, 99]$ and randomly permuting the $M$ machines to determine the machine order for a job's operations. We generate a random validation set of 100 instances of size $20 \times 20$. We perform testing on the widely used benchmark dataset by \citet{taillard1993benchmarks}.

\paragraph{Policy network architecture} We use the recent architecture by \citet{pirnay2024self}, where the operations attend to each other through stacked transformer blocks, switching between different masking schemes. We refer to the appendix of \cite{pirnay2024self} for a detailed architecture description. We use the same setup with six transformer blocks with eight heads and a hidden dimension of 256 in the feedforward network. We note that the downside of the transformer-based architecture is its quadratic complexity with respect to the total number of operations.  

\paragraph{Training} We follow the training scheme of \cite{pirnay2024self}. We train the model with SIL for 450 epochs. In each epoch, we decode solutions to 512 instances of size randomly chosen from $\{15 \times 10, 15 \times 15, 15 \times 20\}$. We use a beam width of $k=64$ and a step size of $s=50$. For the largest size $15 \times 20$, this takes about 5 minutes. As for the routing problems, we set the initial learning rate of the Adam optimizer to 2e-4, clipping gradients to unit norm. In each epoch, we train the model on 1,000 batches consisting of 512 subschedules each. 

\paragraph{Baselines} We compare our method to {\bf (a)} L2D \cite{zhang2020learning} and {\bf (b)} ScheduleNet \cite{park2022schedulenet}, two constructive RL approaches using graph neural networks, {\bf (c)} L2S \cite{zhang2024deep}, a recent impressive deep RL guided improvement heuristic with 500 and 5000 improvement steps, {\bf (d)} SPN \cite{corsini2024self}, a SIL approach which samples sequences with replacement during training, and {\bf (e)} GD \cite{pirnay2024self}, a SIL approach which shares the same network architecture with our model.

\paragraph{Results} We summarize the results in Table~\ref{table:results:jssp}. Our greedy results outperform the baseline greedy results and L2S with 500 steps (comparable runtime) by a wide margin. Notably, we achieve an improvement of $>4\%$ on $30 \times 20$ compared to GD, which uses the same network architecture and already outperforms the other methods. We can further shrink the optimality gap by using our decoding approach as an inference technique (with Top-$p$ sampling with $p=0.9$). Because a solution to a JSSP instance of size $J \times M$ is a sequence of length $J \cdot M$, the results showcase the strength of our sequence decoding method, especially in tasks with a longer planning horizon.

\begin{table*}[!t]
   \caption{JSSP results on the Taillard \cite{taillard1993benchmarks} dataset. Optimal gaps are computed with respect to the best solutions found in the literature by \cite{park2022schedulenet,zhang2024deep,corsini2024self,pirnay2024self}. Results in the bottom row are obtained by applying our proposed decoding method with Top-$p$ sampling, with $p=0.9$}.
   \label{table:results:jssp}
   \begin{center}
   \resizebox{\textwidth}{!}{
     \begin{tabular}{l cc cc cc cc cc cc cc cc}
      & \multicolumn{2}{c}{$15 \times 15$} & \multicolumn{2}{c}{$20 \times 15$} & \multicolumn{2}{c}{$20 \times 20$} & \multicolumn{2}{c}{$30 \times 15$} & \multicolumn{2}{c}{$30 \times 20$} & \multicolumn{2}{c}{$50 \times 15$} & \multicolumn{2}{c}{$50 \times 20$} & \multicolumn{2}{c}{$100 \times 20$} \\
     \Bstrut Method & Gap & Time & Gap & Time & Gap & Time & Gap & Time & Gap & Time & Gap & Time & Gap & Time & Gap & Time \\
     \thickhline
     \Tstrut L2D, greedy \cite{zhang2020learning} & 26.0\% & 0s & 30.0\% & 0s & 31.6\% & 1s & 33.0\% & 1s & 33.6\% & 2s & 22.4\% & 2s & 26.5\% & 4s & 13.6\% & 25s \\
     ScheduleNet, greedy \cite{park2022schedulenet} & 15.3\% & 3s & 19.4\% & 6s & 17.2\% & 11s & 19.1\% & 15s & 23.7\% & 25s & 13.9\% & 50s & 13.5\% & 1.6m & 6.7\% &  7m \\
     L2S, 500 steps \cite{zhang2024deep} & 9.3\% & 9s & 11.6\% & 10s & 12.4\% & 11s & 14.7\% & 12s & 17.5\% & 14s & 11.0\% & 16s & 13.0\% & 23s & 7.9\% & 50s \\
     SPN SIL, greedy \cite{corsini2024self} & 13.8\% & 0s & 15.0\% & 0s & 15.2\% & 0s & 17.1\% & 0s & 18.5\% & 1s & 10.1\% & 1s & 11.6\% & 1s & 5.9\% & 2s \\
     GD SIL, greedy \cite{pirnay2024self} & 9.6\% & 1s & 9.9\% & 1s & 11.1\% & 1s & 9.5\% & 1s & 13.8\% & 2s & 2.7\% & 2s & 6.7\% & 3s & 1.7\% & 28s  \\
     {\bf Ours}, greedy \Bstrut & {\bf 7.7\%} & 1s & {\bf 8.5\%} & 1s & {\bf 8.7\%} & 1s & {\bf 8.4\%} & 1s & {\bf 9.6\%} & 2s & {\bf 2.2\%} & 2s & {\bf 4.9\%} & 3s & {\bf 1.0\%} & 28s  \\
     \hline
     L2S, 5000 steps \cite{zhang2024deep} \Tstrut & 6.2\% & 1.5m & 8.3\% & 1.7m & 9.0\% & 2m & 9.0\% & 2m & 12.6\% & 2.4m & 4.6\% & 2.8m & 6.5\% & 3.8m & 3.0\% & 8.4m \\ 
     {\bf Ours}, $k=64, s=50$ \Bstrut & {\bf 3.0\%} & 10s & {\bf 4.1\%} & 25s & {\bf 4.1\%} & 1m & {\bf 3.9\%} & 90s & {\bf 6.2\%} & 5m & {\bf 0.4\%} & 10m & {\bf 1.7\%} & 30m & {\bf 0.1\%} & 5h\\
     \thickhline
     \end{tabular}
    }
   \end{center}
\end{table*}

\subsection{Sampling comparison}

\begin{figure*}[!th]
      \centering
      \includegraphics[width=0.4\linewidth]{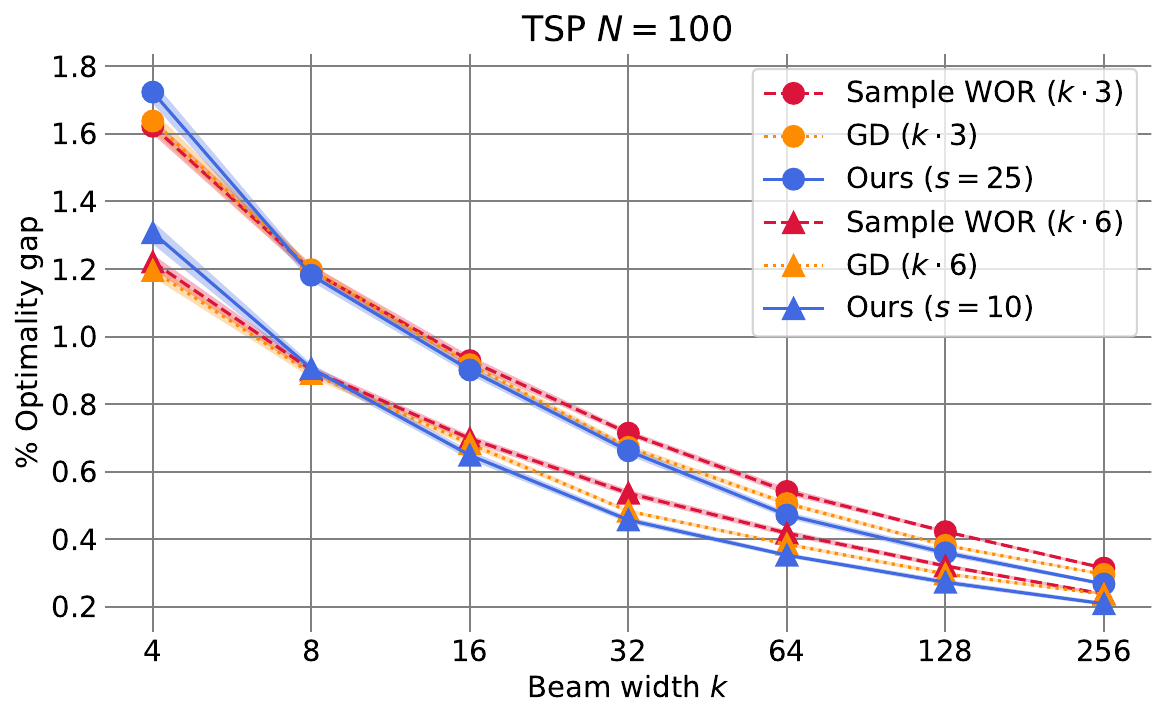}
      \includegraphics[width=0.4\linewidth]{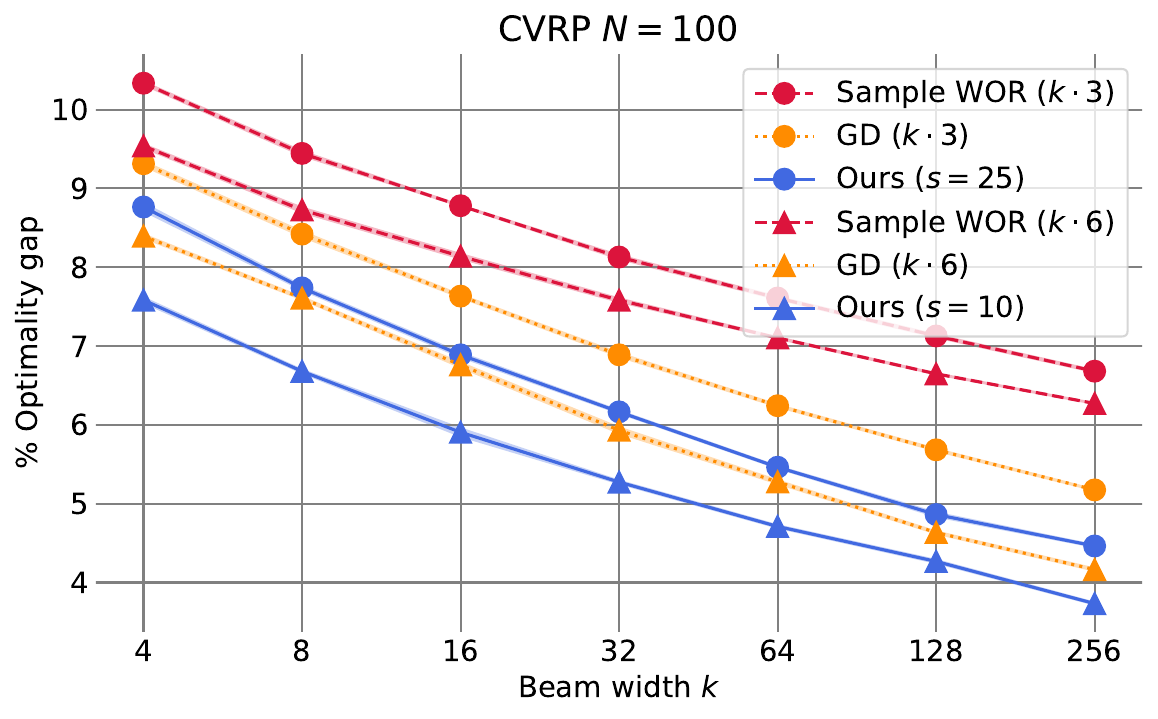}
      \includegraphics[width=0.4\linewidth]{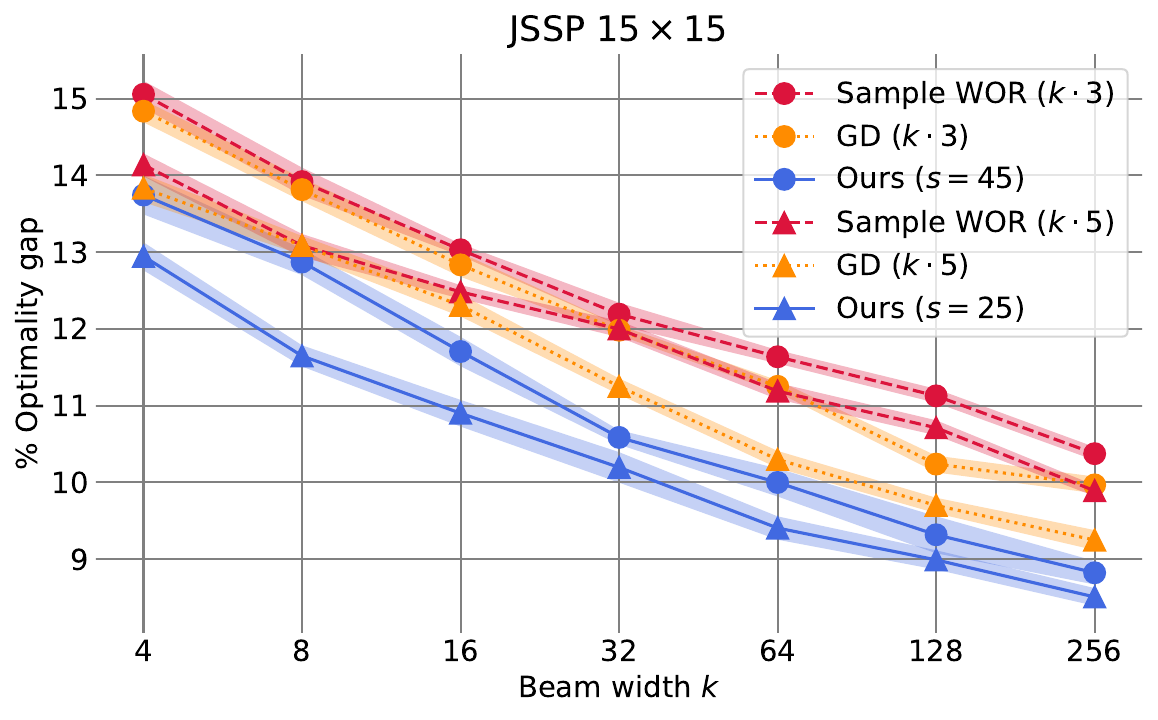}
      \includegraphics[width=0.4\linewidth]{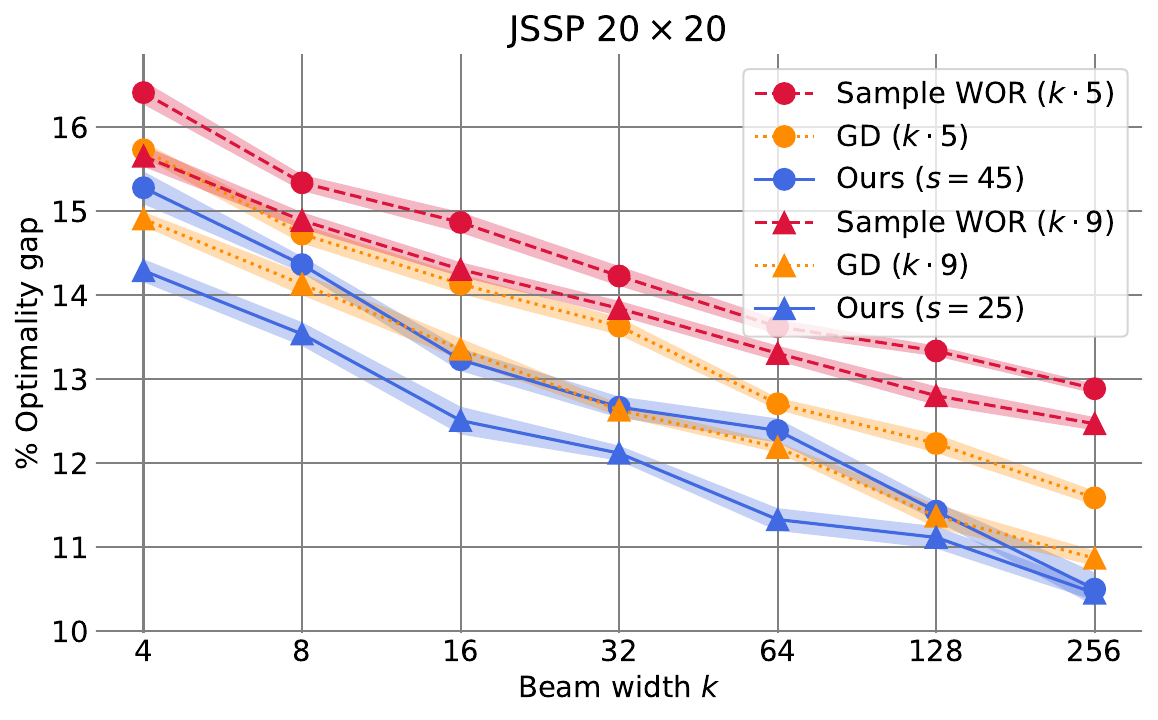}
      \caption{Decoding the policy with our sequence decoding method ('Ours') compared to sampling sequences without replacement with SBS ('Sample WOR') and to the sampling method GD \cite{pirnay2024self}. The number of sequences sampled WOR and with GD are given as multiples of the beam width $k$ to ensure alignment of the computational effort. Points with same marker mean same compute budget. For the routing problems, we average optimality gaps across 100 instances. For the JSSP, the corresponding Taillard benchmark set is used. Sampling for each data point is repeated ten times; shades denote standard errors.}
      \label{fig:sampling_results}
      \vspace{20pt}
\end{figure*}

Our decoding method relies on SBS to sample sequences WOR, which is already an established method to diversify the model output and enhance exploration. SBS is also the basis for the sampling method in GD \cite{pirnay2024self}. Therefore, we compare the quality of the best solution obtained when decoding the policy with our method and sampling sequences WOR (with SBS) and GD. For each considered problem class TSP, CVRP, and JSSP, we take a checkpoint from the middle of the training process when the policy still has room for improvement and exploration is advantageous. We then decode solutions with beam width $k$ and step size $s$ using our method, and also with SBS and GD using the same computational budget. To ensure that we allow SBS and GD at least the same computational budget, we count the number of times we transition from a node in the search tree to a child node. One can show that when $l$ is the length of a complete solution, our decoding method with parameters $k$ and $s$ takes $g(k,s)$ node transitions with
\begin{align}
	g(k,s) = k \cdot \left(tl - \frac{st^2 - st}{2}\right), \text{ where } t = \left\lceil\frac{l}{s}\right\rceil.
\end{align}
Sampling $k$ sequences WOR with SBS takes $k \cdot l$ node transitions. In particular, we allow $k \cdot \left\lceil h(s) \right\rceil$ transitions for SBS and GD, where 
\begin{align}
	h(s) = \frac{g(k,s)}{kl} = \frac{2tl - st^2 + st}{2l}.
\end{align}
For example, for $l=100$, $k=64$ and $s=10$, we have $h(s) = 5.5$, so we grant SBS to sample $6k = 384$ sequences from the root. The same applies to GD, where we sample for $\left\lceil h(s) \right\rceil$ rounds with beam width $k$, using the constants for scaling the advantages in between rounds as reported in \cite{pirnay2024self} (without nucleus sampling).

We show the decoding results in Figure~\ref{fig:sampling_results}. We observe only a small improvement over sampling WOR and GD for the TSP model, as the policy is already confident (< 2\% optimality gap for $k=4$). For the CVRP and the JSSP model, we see a significant improvement of about 1-2\% over sampling WOR, showing that our method can consistently take advantage of the search budget. 

\section{Conclusion} The SIL paradigm, where the neural policy iteratively learns from its own decoded predictions, offers a promising path for NCO to overcome the training complexities and generalization challenges associated with RL methods. However, it requires the construction of a multitude of ever-improving solutions for a substantial number of problem instances during training. Despite this need, there needs to be more guidance apart from standard sequence decoding techniques from natural language processing on {\em how} to effectively build these solutions principled and exploratively. In this paper, we have proposed a novel sequence decoding technique for constructive NCO that is strikingly simple, does not rely on problem specifics, and works particularly well for longer planning horizons. We have achieved this by following a sampled, seemingly good solution for a limited number of steps and, importantly, replanning it by considering previously unseen alternatives. We have demonstrated our method on three prominent CO problems, showing comparable performance to training directly on expert solutions and the ability to surpass existing SIL methods. Notably, we have achieved new state-of-the-art results for NCO on the prominent JSSP Taillard benchmark. Due to its flexibility, our method can in principle also be used in other problem-specific SIL approaches, such as \cite{luo2024self}.



\clearpage
\begin{ack}
This work was funded by the Deutsche Forschungsgemeinschaft (DFG, German Research Foundation) - 466387255 - within the Priority Programme "SPP 2331: Machine Learning in Chemical Engineering". The authors gratefully acknowledge the Competence Center for Digital Agriculture (KoDA) at the University
of Applied Sciences Weihenstephan-Triesdorf for providing additional computational resources.
\end{ack}


\bibliography{m841}

\end{document}